\documentclass[10pt]{article}

\usepackage[letterpaper,margin=1in,top=1in,bottom=1in,headheight=14pt]{geometry}
\usepackage{amsmath,amssymb,amsthm}
\usepackage{graphicx}
\usepackage{booktabs}
\usepackage[dvipsnames,table]{xcolor}
\usepackage{hyperref}
\usepackage{enumitem}
\setlist[itemize]{topsep=2pt,itemsep=2pt,parsep=0pt,partopsep=0pt,leftmargin=1.2em,labelsep=0.5em}
\setlist[enumerate]{topsep=2pt,itemsep=2pt,parsep=0pt,partopsep=0pt,leftmargin=1.5em,labelsep=0.5em}

\usepackage[font=small,labelfont={bf,sf},labelsep=period]{caption}
\usepackage{float}
\usepackage{array}
\usepackage{longtable}
\usepackage{listings}
\usepackage{tabularx}
\usepackage{fancyhdr}
\usepackage{tcolorbox}
\usepackage[protrusion=true,expansion=false]{microtype}
\usepackage{titlesec}
\tcbuselibrary{skins,breakable}
\usepackage{tikz}
\usetikzlibrary{arrows.meta,positioning,calc,shapes.geometric,fit,backgrounds,decorations.pathmorphing}
\usepackage{multicol}

\newcommand{\citelink}[2]{\nocite{#1}#2}

\definecolor{accentteal}{HTML}{1F8A8A}
\definecolor{accentamber}{HTML}{C97A2E}
\definecolor{accentmauve}{HTML}{7B5C8C}
\definecolor{lavrow}{HTML}{EFEAF4}
\definecolor{rulegray}{HTML}{2A2A2A}
\definecolor{softgray}{HTML}{F5F5F5}
\definecolor{hexoorange}{HTML}{F26522}
\definecolor{linkteal}{HTML}{157272}

\hypersetup{
    colorlinks=true,
    linkcolor=linkteal,
    citecolor=linkteal,
    urlcolor=linkteal,
    pdftitle={SIA: Self Improving AI with Harness \& Weight Updates}
}

\titleformat{\section}{\Large\bfseries\sffamily}{\thesection.}{0.6em}{}[{\vspace{-6pt}\color{accentteal}\rule{2.2cm}{1.4pt}}]
\titleformat{\subsection}{\large\bfseries\sffamily}{\thesubsection.}{0.5em}{}
\titleformat{\subsubsection}{\normalsize\bfseries\sffamily}{\thesubsubsection.}{0.4em}{}
\titlespacing*{\section}{0pt}{14pt}{6pt}
\titlespacing*{\subsection}{0pt}{10pt}{4pt}
\titlespacing*{\subsubsection}{0pt}{8pt}{3pt}

\fancypagestyle{plain}{%
  \fancyhf{}%
  \fancyfoot[C]{\small\thepage}%
  \fancyfoot[L]{}%
}

\newcounter{algocount}

\newtcolorbox{keybox}[1][]{%
  enhanced, breakable,
  colback=accentteal!6, colframe=accentteal,
  boxrule=0.6pt, arc=2pt,
  left=10pt, right=10pt, top=6pt, bottom=6pt,
  fonttitle=\bfseries\sffamily\color{white},
  coltitle=white,
  attach boxed title to top left={xshift=8pt,yshift=-7pt},
  boxed title style={colback=accentteal,sharp corners,boxrule=0pt},
  title={#1}
}

\newtcolorbox{rqbox}{%
  colback=lavrow, colframe=accentmauve,
  boxrule=0.5pt, arc=1.5pt,
  left=10pt, right=10pt, top=6pt, bottom=6pt,
}

\makeatletter
\renewcommand{\maketitle}{%
  \thispagestyle{plain}%
  \begin{center}%
    \includegraphics[height=1.2cm]{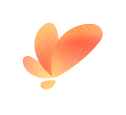}\\[2pt]
    {\small\sffamily\bfseries Hexo Labs}\\[14pt]
    {\LARGE\bfseries\sffamily\@title\par}%
    \vspace{10pt}%
    {\large\@author\par}%
    \vspace{4pt}%
    {\small$^{*}$Equal contribution.\enspace\textperiodcentered\enspace $^{1}$Hexo Labs Palo Alto\enspace\textperiodcentered\enspace$^{2}$Hexo Labs Brussels\enspace\textperiodcentered\enspace$^{3}$Hexo Labs Toronto\enspace\textperiodcentered\enspace$^{4}$University of Oxford}\par
    \vspace{4pt}%
  \end{center}%
  \vspace{4pt}%
}
\makeatother

\title{SIA: Self Improving AI with Harness \& Weight Updates}
\author{Prannay Hebbar$^{*1}$, Yogendra Manawat$^{*1}$, Samuel Verboomen$^{2}$, Alesia Ivanova$^{4}$, Selvam Palanimalai$^{3}$, Kunal Bhatia$^{1}$, Vignesh Baskaran$^{1}$}
\date{}

\begin{document}

\maketitle


\section*{Abstract}

Humans are the bottleneck in building and improving AI. Both the models and the agents that wrap them are written, tuned, and corrected by people. The long-horizon goal of an AI that can figure out how to improve itself remains open. Two largely disjoint research lines attack this bottleneck. The \emph{harness-update school} has a meta-agent rewrite the scaffold of a task-specific agent (its tools, prompts, retry logic, and search procedure) while the model weights are held fixed. The \emph{test-time training school} uses hand-written RL pipelines to update the model's own weights on task feedback while the harness is held fixed. These two silos operate in isolation. We propose \textbf{SIA}, a self-improving loop in which a language-model agent (the \textbf{Feedback-Agent}) updates \emph{both} the harness \emph{and} the weights of a task-specific agent. We evaluate across three contrasting domains: Chinese legal charge classification, low-level GPU kernel optimisation, and single-cell RNA denoising. Combining both levers outperforms scaffold iteration alone on all three benchmarks. SIA-W+H achieves \textbf{25.1\%} over prior SOTA on LawBench, \textbf{12.4\%} faster GPU kernels than prior SOTA (1,017 vs 1,161~$\mu$s), and \textbf{20.4\%} over prior SOTA on denoising. Harness updates make the model agentic, shaping how it searches and acts, while weight updates build the domain intuition that no prompt or scaffold can instil.

\section{Introduction}

\subsection{Humans are the bottleneck.}

Today's progress in AI is rate-limited by humans. The models are designed and post-trained by researchers, and the agents built on top of them are scaffolded, prompted, debugged, and tuned by engineers. The long-horizon goal of the field an AI (model or agent) that can figure out how to improve itself remains open. We treat this paper as one concrete step toward that goal: a system that, given only a task specification and a verifier (both defined in \S3), improves \emph{both} its scaffold and its model weights without further human intervention.

\subsection{Two silos of self-improving AI.}

Research into automated self-improvement has bifurcated into two largely disjoint silos as follows.

\textbf{Silo 1 Harness/scaffold self-improvement.} A meta-agent rewrites the scaffold of the task-specific agent its system prompt, tool-dispatch logic, retry policy, and answer-extraction code across generations, while the underlying language-model weights are held fixed. Recent representatives include the Darwin Gödel Machine (\citelink{zhang2025}{Zhang et al., 2025}), Meta-Harness (\citelink{lee2026}{Lee et al., 2026}), Hyperagents (\citelink{zhang2026}{Zhang et al., 2026}), and the broader line on automated agentic system design (\citelink{hu2024}{Hu et al., 2024}). The recurring empirical observation in this silo is that scaffold edits concentrate on software-engineering hygiene parsing, retries, dispatch and rarely deliver domain-specific reasoning that the base model could not produce given any prompt.

\textbf{Silo 2 Test-time post-training.} A hand-written RL pipeline updates the model's own weights on task feedback at test time, typically with the harness held fixed at a single prompt-and-grader template. Representatives include TTRL (\citelink{zuo2025}{Zuo et al., 2025}), the \emph{Discover} line of test-time training (\citelink{yuksek2026}{Yuksekgonul et al., 2026}), and the surprising-effectiveness-of-TTT result (\citelink{akyurek2024}{Akyürek et al., 2024}). Here the gain comes from internal policy change, but the pipeline that delivers it is engineered by humans and does not adapt to the task structure that a scaffolded agent would expose.

\textbf{The gap.} These two silos operate in isolation. Harness work leaves the model fixed; test-time training leaves the harness fixed.

\subsection{Contributions.}

\begin{itemize}
    \item We propose and evaluate a Feedback-Agent that \textbf{also trains} the task-specific agent's weights, in combination with scaffold updates, to improve performance on arbitrary downstream tasks. The system is task-agnostic: given a task specification and a verifier, it produces both an evolved scaffold and an RL-adapted set of Low-Rank Adaptation (LoRA; \citelink{hu2022lora}{Hu et al., 2022}) weights.
    \item We empirically demonstrate the combined approach across \textbf{three contrasting domains} law (191-class Chinese charge classification; \citelink{fei2023}{Fei et al., 2023}), systems (Triton kernel optimisation on H100; \citelink{novikov2025}{Novikov et al., 2025}), and biology (single-cell RNA denoising; \citelink{vandijk2018}{van Dijk et al., 2018}) and observe \textbf{consistent gains over prior SOTA}: 25.1\% on LawBench (70.1\% vs 45.0\%), 12.4\% faster GPU kernels (1,017 vs 1,161~$\mu$s), and 20.4\% on denoising (0.289 vs 0.240 \texttt{mse\_norm}).
    \item We isolate the \textbf{harness-only} contribution (harness update trajectories across several iterations) and contrast it with the full pipeline (harness + weight updates), demonstrating that weight updates deliver gains beyond what the harness alone achieves.
\end{itemize}

\subsection{Roadmap.}

\S2 states the research questions the paper answers and maps each to a later section. \S3 defines the technical vocabulary. \S4 places SIA in the landscape of self-improving and test-time-training work. \S5 describes the configurable-loop method. \S6 presents the per-task results and ablations. \S7 discusses what each lever changes. \S8 and \S9 close with limitations and future work.

\begin{figure}[H]
\centering
\includegraphics[width=\linewidth]{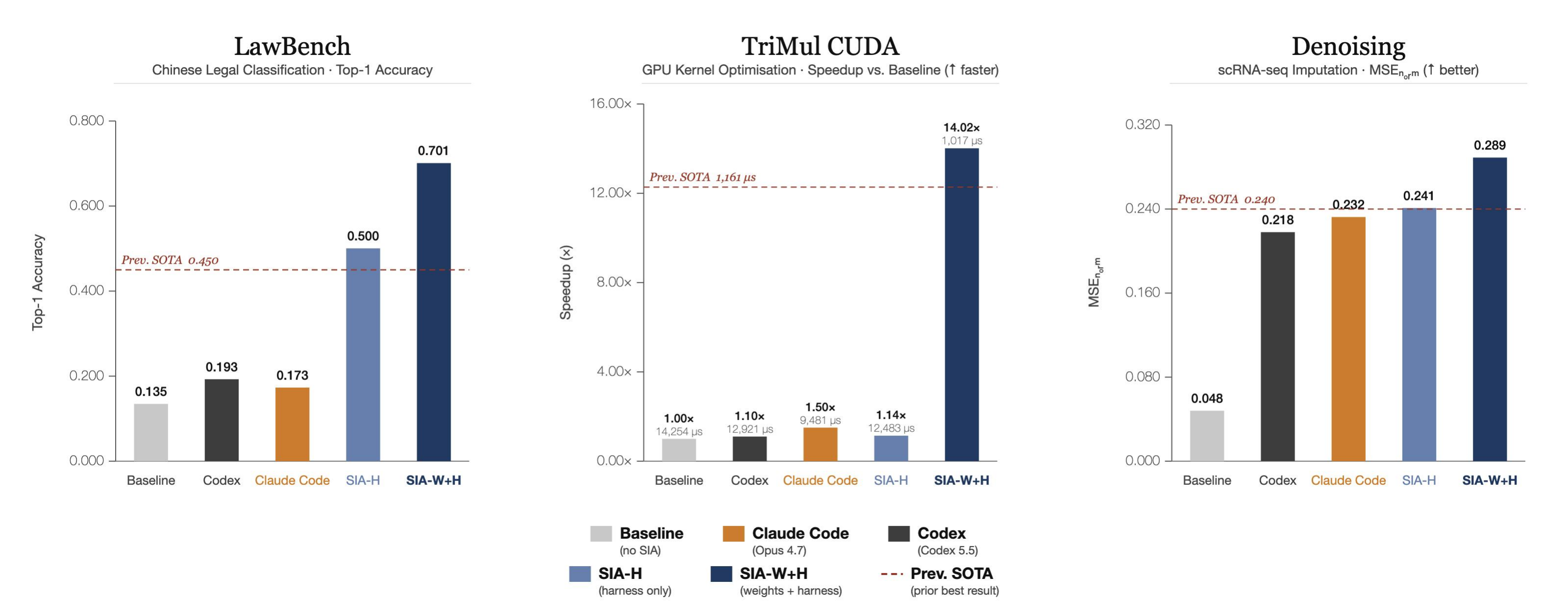}
\caption{\textbf{SIA across three diverse tasks.} Each panel compares three operating points: Baseline (first generation, no SIA), SIA-H (harness updates only), and SIA-W+H (harness + weight updates), on LawBench Top-1 accuracy, TriMul CUDA speedup, and scRNA-seq denoising \texttt{mse\_norm}. The dashed line marks the previous state-of-the-art. SIA-W+H strictly outperforms SIA-H on all three tasks.}
\label{fig:1}
\end{figure}

\section{Research Questions}

This paper is organised around two research questions. Each is answered by a specific later section.

\begin{rqbox}
\begin{itemize}
    \item \textbf{RQ1 Overall thesis.} We first ask how much harness iteration alone improves a task-specific agent when model weights are held fixed. We then ask whether running both levers together (iteratively updating the harness \emph{and} the model weights in a single loop) pushes past that harness-only ceiling. Does the combined approach outperform scaffold iteration alone, and does this hold across contrasting domains?
    \item \textbf{RQ2 Mechanism: what does each lever change?} Do weight updates surface domain knowledge that no scaffold edit reaches, and does harness iteration produce qualitatively different (external infrastructure) changes?
\end{itemize}
\end{rqbox}

\section{Background and Preliminaries}

\subsection{Agent and its components.}

A \textbf{task-specific agent} is a program that takes a task instance and produces an answer. We decompose it into:

\begin{itemize}
    \item \textbf{LLM.} The underlying language model with weights $\theta$. We use \texttt{openai/gpt-oss-120b} as the base model throughout.
    \item \textbf{System prompt.} Fixed text prepended to every model call that frames the task.
    \item \textbf{Tool-dispatch logic.} Python code that parses model tool-call outputs and routes them to handlers (file I/O, code execution, dataset lookup, grader calls).
    \item \textbf{Answer extraction.} Code that converts a model response (typically a structured trailing block) into a benchmark-formatted prediction.
    \item \textbf{Grader.} The deterministic verifier the orchestrator invokes to compute the per-instance reward.
\end{itemize}

We call the fixed, non-weight component of the agent the \textbf{scaffold} (equivalently, \textbf{harness}) throughout. It is the union of the system prompt, tool-dispatch logic, answer extraction, and any supporting infrastructure, every part of the agent that is fixed code rather than model output.

\subsection{Meta-agent vs. task-specific agent.}\label{sec:components}

A \textbf{meta-agent} is an LLM call whose output is itself an agent. SIA uses two meta-agents:

\begin{itemize}
    \item \textbf{Meta-Agent ($\mathcal{M}$).} Generates the initial scaffold $A_1$ from the task specification $\mathcal{U}$ and any reference implementations $\mathcal{R}$ supplied with the benchmark:
    $$A_1 = \mathcal{M}(\mathcal{U},\, \mathcal{R}).$$
    \item \textbf{Feedback-Agent ($\mathcal{F}$).} Reads the previous generation's scaffold $A_g$, its execution trajectory $\tau_g$, and performance metrics $\mathcal{E}_g$, and synthesises an improved scaffold:
    $$A_{g+1} = \mathcal{F}(A_g,\, \tau_g,\, \mathcal{E}_g,\, \mathcal{U}).$$
\end{itemize}
The \textbf{task-specific agent} is the scaffold $A_g$ at generation $g$ that actually executes against the evaluation dataset.

\subsection{Trajectory and feedback loop.}

Unlike systems that condition improvement on aggregate metrics alone, $\mathcal{F}$ receives the full \textbf{trajectory} $\tau_g$, the complete structured execution log from running $A_g$ against $\mathcal{D}$ (the evaluation dataset): every prompt, model response, tool call, tool result, and extracted answer for every task instance. This allows $\mathcal{F}$ to diagnose specific failure modes rather than react to summary statistics.

Each generation $g$ follows a three-phase protocol:

\begin{enumerate}
    \item \textbf{Execution.} $A_g$ runs on $\mathcal{D}$ inside a sandbox: read-only access to the dataset directory, read/write access to a working directory. The trajectory $\tau_g$ is captured.
    \item \textbf{Analysis.} $\mathcal{F}$ receives $A_g$'s source code, $\tau_g$, the metrics $\mathcal{E}_g$, and optionally sample task descriptions used to discourage single-instance overfitting.
    \item \textbf{Improvement.} $\mathcal{F}$ emits two artefacts: an \emph{improvement report} (prose analysis and the proposed changes) and the \emph{next-generation agent} $A_{g+1}$.
\end{enumerate}

\subsection{Symbol table.}

\begin{table}[H]
\centering
\begin{tabular}{ll}
\toprule
\textbf{Symbol} & \textbf{Meaning} \\
\midrule
$g$ & Generation index \\
$G_{\max}$ & Maximum number of generations \\
$A_g$ & Agent scaffold at generation $g$ \\
$\mathcal{D}$ & Evaluation dataset \\
$\mathcal{U}$ & Task specification (benchmark description + sample instances) \\
$\mathcal{E}_g$ & Performance metrics and error logs at generation $g$ \\
$\tau_g$ & Execution trajectory at generation $g$ \\
$\mathcal{F}$ & Feedback agent \\
$G$ & Number of rollouts per state during RL training \\
$\pi_\theta$ & Current policy (model with trainable weights $\theta$) \\
$\pi_{\theta_0}$ & Frozen reference policy (base model) \\
$s$ & Initial state (task prompt) \\
$a$ & Action (model-generated response / rollout) \\
$V(s,a)$ & Task reward for action $a$ given state $s$ \\
\bottomrule
\end{tabular}
\end{table}

\section{Related Work}

We survey each silo, characterise the specific gap SIA addresses, and summarise the landscape in a comparison table.

\subsection{Harness / scaffold self-improvement.}

\begin{itemize}
    \item \textbf{Darwin Gödel Machine (\citelink{zhang2025}{Zhang et al., 2025}).} Evolutionary search over agent source code: a population of agents proposes and evaluates code mutations to themselves, with the highest-fitness variants surviving. The model is fixed.
    \item \textbf{Meta-Harness (\citelink{lee2026}{Lee et al., 2026}).} LLM-driven harness mutation with end-to-end optimisation of the harness graph. SIA's harness update step is closest to Meta-Harness in spirit; the difference is that we follow harness convergence with weight updates rather than further mutation.
    \item \textbf{Hyperagents (\citelink{zhang2026}{Zhang et al., 2026}).} The closest concurrent work. Hyperagents allows the \emph{meta-mechanism} itself the rules by which the meta-agent edits the task-specific agent to be editable, not just the task-specific agent. The agent and the agent-improver coevolve. The distinction from SIA is the lever: Hyperagents adds expressivity to scaffold edits but leaves the model weights fixed; SIA adds a second, weight-based lever.
    \item \textbf{AI Scientist (\citelink{lu2024}{Lu et al., 2024}).} A full research-pipeline meta-agent that proposes hypotheses, runs experiments, writes papers. The agent's outputs are research artefacts, not modified scaffolds; the scaffold is held fixed across runs.
    \item \textbf{Automated design of agentic systems (\citelink{hu2024}{Hu et al., 2024}).} Meta-search over compositions of building blocks (sub-agents, tools, prompts). Model fixed.
    \item \textbf{AutoResearcher (\citelink{karpathy2026}{Karpathy, 2026}).} A static scaffold for autonomous ML experimentation: the agent proposes and runs experiment configurations, but the agent architecture itself does not change across iterations. \end{itemize}

\subsection{Test-time training and test-time RL.}

\begin{itemize}
    \item \textbf{Learning to discover at test time (\citelink{yuksek2026}{Yuksekgonul et al., 2026}).} The objective we use in training update steps. Trains weights at test time using rollouts under an entropic-utility objective; SIA reuses this loss and the LoRA-based training stack.
    \item \textbf{Surprising effectiveness of TTT (\citelink{akyurek2024}{Akyürek et al., 2024}).} Empirical demonstration that per-task gradient adaptation at test time substantially improves few-shot performance. Establishes the TTT-as-adaptation framing.
    \item \textbf{TTRL (\citelink{zuo2025}{Zuo et al., 2025}).} RL on unlabelled test data using majority-vote-derived pseudo-rewards. The setting is single-prompt, single-response; there is no scaffold and no per-instance verifier. SIA differs in that the reward is a deterministic task verifier and the rollout is scaffolded.
    \item \textbf{STaR (\citelink{zelikman2022}{Zelikman et al., 2022}); Self-Refine (\citelink{madaan2023}{Madaan et al., 2023}); Reflexion (\citelink{shinn2023}{Shinn et al., 2023}).} Earlier self-improvement loops that bootstrap reasoning traces or use verbal critique. STaR fine-tunes the model on self-generated rationales (a supervised weight update); Self-Refine and Reflexion operate purely at inference time with no weight updates.
    \item \textbf{Self-play fine-tuning (\citelink{chen2024}{Chen et al., 2024}).} Iterative fine-tuning where the model's own outputs serve as training signal. The training pipeline is hand-written; the scaffold is fixed.
    \item \textbf{EUREKA (\citelink{ma2024eureka}{Ma et al., 2023}).} An LLM generates reward functions (a scaffold-side change), which are then used to train RL policies (a weight-side change). The two components interact, but the reward-function generator is not itself updated by the trained policy, the loop is one-directional rather than co-evolutionary. SIA differs in that the Feedback-Agent dynamically selects between scaffold and weight updates in a closed feedback loop, with each update type informed by trajectories produced under the current state of both components.
\end{itemize}

\subsection{RL and agent training infrastructure.}

Across all training runs, we use \texttt{gpt-oss-120b} with LoRA rank~32 as the base model and adapter configuration. Weight updates are executed on H100 GPUs via \textbf{Modal}, our RL training platform, which handles rollout generation, reward assignment, and gradient updates within a single managed pipeline. SIA builds on existing training frameworks; the Feedback-Agent composes these infrastructure components under its control, treating weight updates as one of two selectable actions alongside scaffold rewriting. Related infrastructure includes \textbf{verl/HybridFlow} (\citelink{sheng2024}{Sheng et al., 2024}) for flexible RLHF, \textbf{SkyRL} (\citelink{cao2025}{Cao et al., 2025}) for long-horizon agent training, \textbf{LLaMA-Factory} (\citelink{zheng2024}{Zheng et al., 2024}) for unified post-training, and \textbf{Axolotl} for streamlined fine-tuning configurations.

\subsection{Comparison table.}

\begin{table}[H]
\centering
\captionsetup{font=small,skip=6pt,labelfont={bf,sf},labelsep=period}
\caption{\textbf{Comparison of self-improving / automated agents along two axes.} Does the system edit the harness? Does it edit the model weights?}
\footnotesize
\renewcommand{\arraystretch}{1.25}
\setlength{\tabcolsep}{5pt}
\rowcolors{3}{softgray}{white}
\begin{tabular}{@{}lcc@{}}
\toprule
\textbf{Agent} & \textbf{Edits harness} & \textbf{Edits weights} \\
\midrule
\rowcolor{lavrow}\textbf{SIA (ours)} & \textbf{Yes} & \textbf{Yes} \\
Hyperagents (\citelink{zhang2026}{Zhang et al., 2026}) & Yes & No \\
Darwin Gödel Machine (\citelink{zhang2025}{Zhang et al., 2025}) & Yes & No \\
Meta-Harness (\citelink{lee2026}{Lee et al., 2026}) & Yes & No \\
AI Scientist (\citelink{lu2024}{Lu et al., 2024}) & Partial & No \\
Automated agentic system design (\citelink{hu2024}{Hu et al., 2024}) & Yes & No \\
AutoResearcher (\citelink{karpathy2026}{Karpathy, 2026}) & No & No \\
TTRL (\citelink{zuo2025}{Zuo et al., 2025}) & No & Yes \\
Discover-TTT (\citelink{yuksek2026}{Yuksekgonul et al., 2026}; \citelink{akyurek2024}{Akyürek et al., 2024}) & No & Yes \\
EUREKA (\citelink{ma2024eureka}{Ma et al., 2023}) & Partial & Yes \\
FunSearch (\citelink{romera2024}{Romera-Paredes et al., 2024}) & Partial & No \\
Voyager (\citelink{wang2023}{Wang et al., 2023}) & Yes & No \\
Self-Refine (\citelink{madaan2023}{Madaan et al., 2023}) / Reflexion (\citelink{shinn2023}{Shinn et al., 2023}) & Partial & No \\
STaR (\citelink{zelikman2022}{Zelikman et al., 2022}) & No & Yes \\
ReAct (\citelink{yao2022}{Yao et al., 2022}) & No & No \\
\bottomrule
\end{tabular}
\end{table}

SIA is, to our knowledge, the only entry that updates both the scaffold and the weights in a single self-improving loop.

\section{Method}

\subsection{Overview.}

SIA is a configurable loop driven by three LLM components: a Meta-Agent, a Task-Specific Agent, and a Feedback-Agent. The Meta-Agent initialises the task-specific agent's scaffold. After each execution, the Feedback-Agent observes the trajectory and performance, then dynamically selects, at each step, between two complementary actions: a \textbf{harness update} (scaffold evolution with weights fixed) or a \textbf{training algorithm update} (weight update via an RL method of the Feedback-Agent's choosing, with the scaffold fixed). The choice of action, and the choice of training algorithm when a weight update is selected, are conditioned on task type and observed reward dynamics. \textbf{Harness Update Phase} and \textbf{Weight Update Phase} are soft labels for these two action types, not rigid sequential stages.

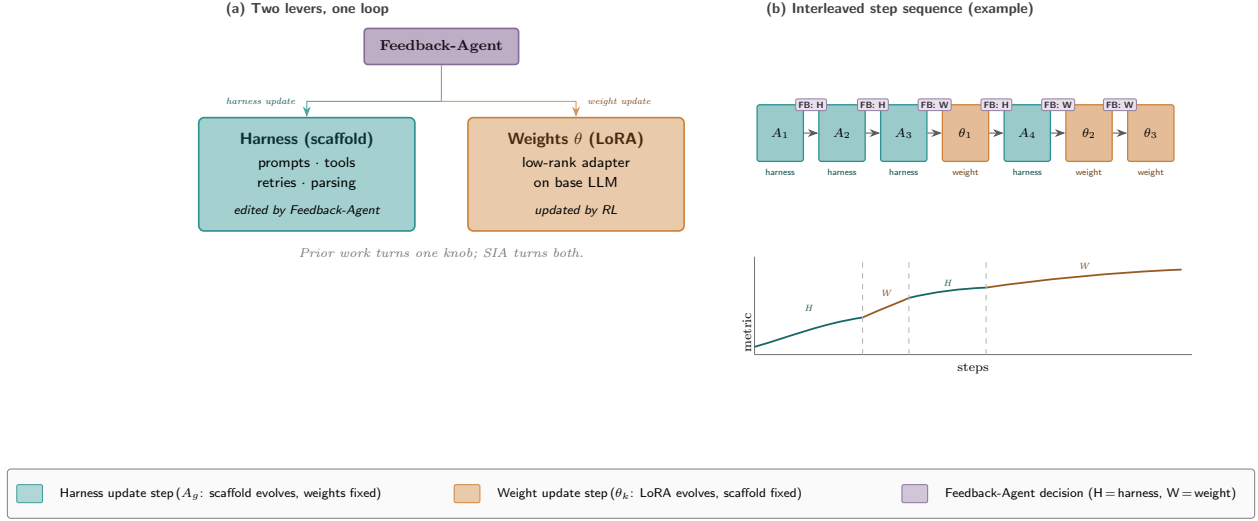
\begin{figure}[H]
\centering
\resizebox{\linewidth}{!}{%
\begin{tikzpicture}[font=\small\sffamily,
    leverbox/.style={draw, rounded corners=3pt, line width=0.6pt, minimum width=42mm, minimum height=22mm, align=center, inner sep=6pt},
    miniH/.style={draw=accentteal!95, fill=accentteal!45, rounded corners=1.5pt, line width=0.7pt, minimum width=9mm, minimum height=11mm, font=\scriptsize, align=center},
    miniW/.style={draw=accentamber!95, fill=accentamber!50, rounded corners=1.5pt, line width=0.7pt, minimum width=9mm, minimum height=11mm, font=\scriptsize, align=center},
    fblabel/.style={draw=accentmauve!80, fill=accentmauve!18, rounded corners=1pt, font=\tiny\sffamily\bfseries, text=accentmauve!40!black, inner sep=1.5pt, line width=0.5pt},
    arr/.style={-{Stealth[length=2mm]}, line width=0.5pt, draw=black!65}]

\node[anchor=west, font=\footnotesize\bfseries\sffamily, color=black!75] (Atitle) at (0,5.2) {(a)\,\,Two levers, one loop};

\node[leverbox, fill=accentteal!40, draw=accentteal!95, line width=0.9pt] (harness) at (1.7,2.0) {%
  \textbf{\color{accentteal!35!black}Harness (scaffold)}\\[2pt]
  \footnotesize prompts $\cdot$ tools\\
  \footnotesize retries $\cdot$ parsing\\[4pt]
  \scriptsize\itshape edited by Feedback-Agent};

\node[leverbox, fill=accentamber!40, draw=accentamber!95, line width=0.9pt, right=10mm of harness] (weights) {%
  \textbf{\color{accentamber!35!black}Weights $\theta$ (LoRA)}\\[2pt]
  \footnotesize low-rank adapter\\
  \footnotesize on base LLM\\[4pt]
  \scriptsize\itshape updated by RL};

\node[draw=accentmauve!90, fill=accentmauve!50, rounded corners=2pt, font=\footnotesize\bfseries,
      text=accentmauve!20!black, line width=0.9pt,
      above=10mm of $(harness.north)!0.5!(weights.north)$,
      minimum width=30mm, minimum height=7mm] (fbA) {Feedback-Agent};

\coordinate (forkA) at ($(harness.north)!0.5!(weights.north)+(0,3mm)$);
\draw[draw=accentmauve!70, line width=0.6pt] (fbA.south) -- (forkA);
\draw[arr, draw=accentteal!80] (forkA) -| (harness.north)
    node[midway, left, font=\tiny\itshape, color=accentteal!60!black, xshift=-1mm] {harness update};
\draw[arr, draw=accentamber!80] (forkA) -| (weights.north)
    node[midway, right, font=\tiny\itshape, color=accentamber!60!black, xshift=1mm] {weight update};

\node[font=\scriptsize\itshape, color=black!55, anchor=north] at ($(harness.south)!0.5!(weights.south)+(0,-2mm)$) {Prior work turns one knob; SIA turns both.};

\node[anchor=west, font=\footnotesize\bfseries\sffamily, color=black!75] (Btitle) at (10.5,5.2) {(b)\,\,Interleaved step sequence (example)};

\node[miniH] (St1) at (10.9, 2.8) {$A_1$};
\node[miniH] (St2) at (12.1, 2.8) {$A_2$};
\node[miniH] (St3) at (13.3, 2.8) {$A_3$};
\node[miniW] (St4) at (14.5, 2.8) {$\theta_1$};
\node[miniH] (St5) at (15.7, 2.8) {$A_4$};
\node[miniW] (St6) at (16.9, 2.8) {$\theta_2$};
\node[miniW] (St7) at (18.1, 2.8) {$\theta_3$};

\foreach \a/\b in {St1/St2, St2/St3, St3/St4, St4/St5, St5/St6, St6/St7}{
    \draw[arr] (\a.east) -- (\b.west);
}

\node[fblabel] at ($(St1.east)!0.5!(St2.west)+(0,5.5mm)$) {FB: H};
\node[fblabel] at ($(St2.east)!0.5!(St3.west)+(0,5.5mm)$) {FB: H};
\node[fblabel] at ($(St3.east)!0.5!(St4.west)+(0,5.5mm)$) {FB: W};
\node[fblabel] at ($(St4.east)!0.5!(St5.west)+(0,5.5mm)$) {FB: H};
\node[fblabel] at ($(St5.east)!0.5!(St6.west)+(0,5.5mm)$) {FB: W};
\node[fblabel] at ($(St6.east)!0.5!(St7.west)+(0,5.5mm)$) {FB: W};

\node[font=\tiny\sffamily, color=accentteal!55!black, anchor=north] at (St1.south) {harness};
\node[font=\tiny\sffamily, color=accentteal!55!black, anchor=north] at (St2.south) {harness};
\node[font=\tiny\sffamily, color=accentteal!55!black, anchor=north] at (St3.south) {harness};
\node[font=\tiny\sffamily, color=accentamber!55!black, anchor=north] at (St4.south) {weight};
\node[font=\tiny\sffamily, color=accentteal!55!black, anchor=north] at (St5.south) {harness};
\node[font=\tiny\sffamily, color=accentamber!55!black, anchor=north] at (St6.south) {weight};
\node[font=\tiny\sffamily, color=accentamber!55!black, anchor=north] at (St7.south) {weight};

\begin{scope}[shift={(10.4,-1.5)}]
\draw[draw=black!55, line width=0.5pt] (0,0) -- (8.5,0);
\draw[draw=black!55, line width=0.5pt] (0,0) -- (0,1.9);
\node[font=\scriptsize, anchor=east, rotate=90] at (-0.15,0.95) {metric};
\node[font=\scriptsize, anchor=north] at (4.25,-0.05) {steps};
\draw[line width=1pt, draw=accentteal!75!black]
    (0.0,0.15) .. controls (0.7,0.35) and (1.3,0.60) .. (2.1,0.72);
\draw[line width=1pt, draw=accentamber!75!black]
    (2.1,0.72) .. controls (2.5,0.90) and (2.9,1.05) .. (3.0,1.10);
\draw[line width=1pt, draw=accentteal!75!black]
    (3.0,1.10) .. controls (3.5,1.22) and (4.0,1.28) .. (4.5,1.30);
\draw[line width=1pt, draw=accentamber!75!black]
    (4.5,1.30) .. controls (5.5,1.45) and (7.0,1.60) .. (8.3,1.65);
\foreach \x in {2.1, 3.0, 4.5}{
    \draw[draw=black!30, line width=0.4pt, dashed] (\x, 0) -- (\x, 1.85);
}
\node[font=\tiny\itshape, color=accentteal!60!black] at (1.05, 0.90) {H};
\node[font=\tiny\itshape, color=accentamber!60!black] at (2.55, 1.20) {W};
\node[font=\tiny\itshape, color=accentteal!60!black] at (3.75, 1.40) {H};
\node[font=\tiny\itshape, color=accentamber!60!black] at (6.40, 1.72) {W};
\end{scope}

\begin{scope}[shift={(0,-4.2)}]
  \node[draw=black!50, rounded corners=2pt, line width=0.4pt, inner sep=5pt, fill=black!2] at (8.0,0) {%
    \begin{tikzpicture}[font=\scriptsize\sffamily]
      \node[draw=accentteal!85, fill=accentteal!35, rounded corners=1pt, minimum width=5mm, minimum height=3.5mm, inner sep=0pt] (lA) at (0,0) {};
      \node[anchor=west] at ($(lA.east)+(1.5mm,0)$) {Harness update step\,($A_g$: scaffold evolves, weights fixed)};
      \node[draw=accentamber!90, fill=accentamber!40, rounded corners=1pt, minimum width=5mm, minimum height=3.5mm, inner sep=0pt] (lT) at (8.5,0) {};
      \node[anchor=west] at ($(lT.east)+(1.5mm,0)$) {Weight update step\,($\theta_k$: LoRA evolves, scaffold fixed)};
      \node[draw=accentmauve!85, fill=accentmauve!35, rounded corners=1pt, minimum width=5mm, minimum height=3.5mm, inner sep=0pt] (lF) at (17.2,0) {};
      \node[anchor=west] at ($(lF.east)+(1.5mm,0)$) {Feedback-Agent decision (H\,=\,harness, W\,=\,weight)};
    \end{tikzpicture}%
  };
\end{scope}

\end{tikzpicture}}
\caption{\textbf{Conceptual view of SIA.} (a) Two complementary levers (a textual scaffold and a LoRA adapter). After each execution, the Feedback-Agent (mauve) selects the next action: a harness update (teal) or a weight update (amber). The two levers are interleaved freely, not locked into sequential phases. (b) An example 7-step sequence showing the Feedback-Agent alternating between harness and weight updates. Each \textbf{FB:H}/\textbf{FB:W} badge marks one decision. The metric curve rises from both types of step, with harness updates (teal segments) and weight updates (amber segments) each contributing distinct gains.}
\label{fig:concept}
\end{figure}

\begin{figure}[H]
\centering
\scalebox{0.82}{%
\begin{tikzpicture}[
    font=\small\sffamily,
    mainbox/.style={draw, rounded corners=3pt, minimum height=11mm, minimum width=42mm,
                    align=center, line width=0.8pt, inner sep=6pt},
    metabox/.style={mainbox, fill=black!22, draw=black!65,
                    text=black!85},
    targetbox/.style={mainbox, fill=accentteal!40, draw=accentteal!90,
                      text=accentteal!20!black},
    fbbox/.style={mainbox, fill=accentmauve!50, draw=accentmauve!90,
                  text=accentmauve!20!black},
    envbox/.style={mainbox, fill=black!7, draw=black!45, text=black},
    iobox/.style={draw, rounded corners=2pt, fill=black!7, draw=black!45,
                  minimum height=9mm, minimum width=32mm, align=center, line width=0.6pt},
    arr/.style={-{Stealth[length=2.5mm]}, line width=0.8pt, draw=black!70},
]

\node[iobox] (task) {Task spec $\mathcal{U}$};
\node[iobox, below=10mm of task] (verif) {Verifier $V$};

\node[metabox, right=26mm of task, yshift=-5mm] (meta) {Meta-Agent};

\node[targetbox, below=18mm of meta] (target) {Task-Specific Agent};

\node[envbox, below=18mm of target] (env) {Environment};

\node[fbbox, right=26mm of env] (fb) {Feedback-Agent};

\coordinate (join) at ($(meta.west)+(-9mm,0)$);
\draw[line width=0.7pt, draw=black!60] (task.east) -| (join);
\draw[line width=0.7pt, draw=black!60] (verif.east) -| (join);
\draw[arr] (join) -- (meta.west);

\draw[arr] (meta.south) -- (target.north);

\draw[arr] (target.south) -- (env.north);

\draw[arr] (env.east) -- (fb.west);

\draw[arr] (fb.north)
    -- node[right=3mm, font=\small\itshape, align=left]
           {update harness\\or weights}
    (fb.north |- target.east)
    -- (target.east);

\node[draw=black!40, rounded corners=3pt, line width=0.5pt, fill=black!2,
      inner sep=8pt, below=14mm of env, anchor=north] (legend) {%
  \begin{tikzpicture}[font=\footnotesize\sffamily, node distance=4mm]
    \node[draw=black!65, fill=black!22, rounded corners=2pt,
          minimum width=6mm, minimum height=4mm, inner sep=0pt] (l1) at (0,0) {};
    \node[anchor=west] (l1t) at ($(l1.east)+(2mm,0)$) {Meta-Agent: initialises the scaffold};

    \node[draw=accentteal!90, fill=accentteal!40, rounded corners=2pt,
          minimum width=6mm, minimum height=4mm, inner sep=0pt,
          right=18mm of l1t.east, anchor=west] (l2) {};
    \node[anchor=west] at ($(l2.east)+(2mm,0)$) {Task-Specific Agent: executes the task};

    \node[draw=accentmauve!90, fill=accentmauve!50, rounded corners=2pt,
          minimum width=6mm, minimum height=4mm, inner sep=0pt,
          below=5mm of l1] (l3) {};
    \node[anchor=west] (l3t) at ($(l3.east)+(2mm,0)$) {Feedback-Agent: selects next action};

    \node[draw=black!45, fill=black!7, rounded corners=2pt,
          minimum width=6mm, minimum height=4mm, inner sep=0pt,
          right=18mm of l3t.east, anchor=west] (l4) {};
    \node[anchor=west] at ($(l4.east)+(2mm,0)$) {Environment \& Inputs: fixed context};
  \end{tikzpicture}%
};

\end{tikzpicture}}
\caption{\textbf{SIA system architecture.} The Meta-Agent initialises a scaffold from the task specification $\mathcal{U}$ and verifier $V$. The Task-Specific Agent executes inside the Environment, producing a trajectory; the Feedback-Agent analyses the trajectory and selects the next action, either synthesising an improved scaffold (harness update) or triggering a weight update, then feeds the result back to the Task-Specific Agent. The loop repeats until the step budget is exhausted.}
\label{fig:2}
\end{figure}
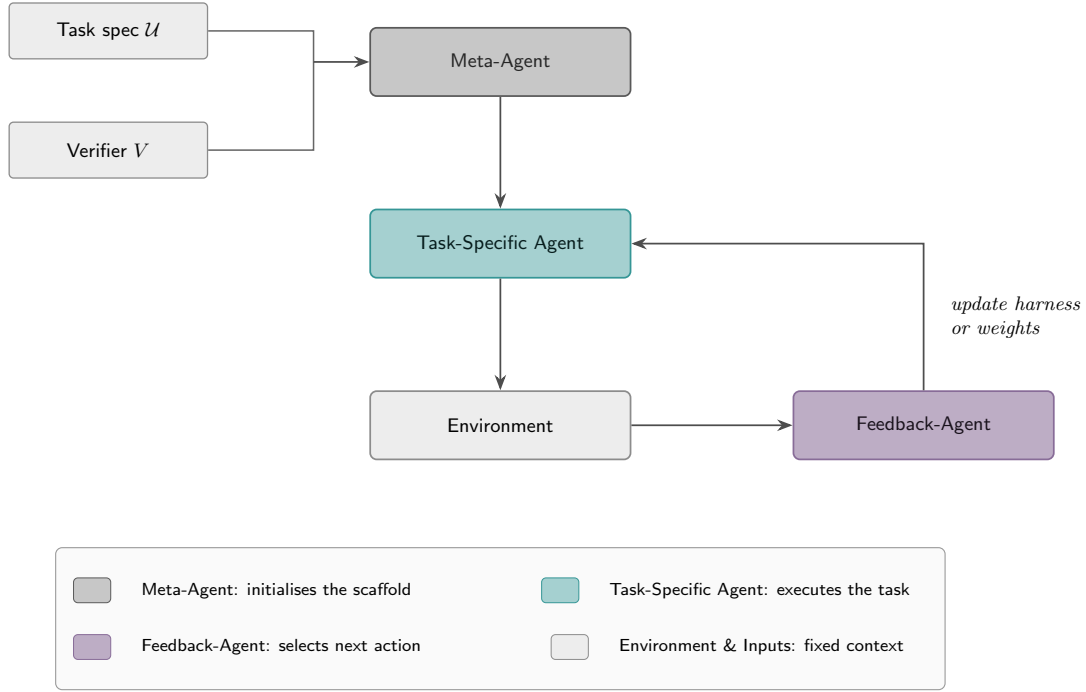

\subsection{System components.}

SIA's three components (Meta-Agent, Task-Specific Agent, Feedback-Agent) are defined in \S\ref{sec:components}; implementation details follow here. Across all experiments, the Meta-Agent and Feedback-Agent use \texttt{Claude Sonnet 4.6}; the task-specific agent uses \texttt{gpt-oss-120b} (harness steps) or an RL-adapted checkpoint thereof (training steps).

\subsection{Harness updates.}\label{sec:harness-update}

When the Feedback-Agent selects a harness update, the loop runs one scaffold evolution step. Each such step follows the per-step protocol (Execution $\to$ Analysis $\to$ Improvement). Rollouts are produced by the current model $\pi_\theta$ (base or RL-adapted); the model weights $\theta$ are held fixed during this step and only the scaffold $A_g$ changes. The recurrence is

$$A_{g+1} = \mathcal{F}(A_g,\, \tau_g(\pi_\theta),\, \mathcal{E}_g,\, \mathcal{U}),$$

where $\tau_g(\pi_\theta)$ denotes trajectories collected by executing scaffold $A_g$ with model $\pi_\theta$.

\textbf{Sample-task regularisation.} The Meta-Agent is conditioned on a diverse set of task specifications during scaffold generation, which mitigates overfitting the initial scaffold to a single benchmark instance.

\section{Experiments}\label{sec:experiments}

We evaluate SIA on three contrasting tasks spanning law, systems, and biology. These benchmarks are commonly used to evaluate other self-improving AI systems; we run on them specifically to enable direct comparison against prior work.

\subsection{Setup.}

\begin{table}[H]
\centering
\captionsetup{font=small,skip=6pt,labelfont={bf,sf},labelsep=period}
\caption{\textbf{Per-task evaluation setup.}}
\footnotesize
\renewcommand{\arraystretch}{1.5}
\rowcolors{2}{softgray}{white}
\begin{tabularx}{\linewidth}{l|X|X|X|X|X}
\toprule
\rowcolor{white}
\textbf{Task} & \textbf{Domain} & \textbf{Train / Test} & \textbf{Metric} & \textbf{Previous SOTA} & \textbf{Verifier} \\
\midrule
LawBench (191-class) & Chinese legal & 5,332 / 913 & top-1 accuracy & 0.450 & held-out test-split grader \\
AlphaEvolve TriMul & Low-level & n/a / fixed input shape & $\text{score} = 1500/\text{runtime}$ (higher = faster) & 1.292 & H100 timing \\
MAGIC scRNA-seq Denoising & Single-cell & n/a / pancreas scRNA-seq & \texttt{mse\_norm} ($\in [0,1]$, higher = better) & 0.24 & MAGIC reference against ground truth \\
\bottomrule
\end{tabularx}
\end{table}

All weight update steps adapt \texttt{gpt-oss-120b} via LoRA (rank $r=32$, learning rate $4\times10^{-5}$).

\subsection{Baselines.}

Because harness update steps start from a meta-agent-initialised scaffold around \texttt{gpt-oss-120b} and run against the same verifier we report, \textbf{the initial score is, by construction, \texttt{gpt-oss-120b} filtered through the Meta-Agent's initial scaffold $A_1$ (a task-specific system prompt, single-tool dispatch loop, and output parser generated once from the benchmark specification before any Feedback-Agent iteration begins)}. The harness update trajectory then traces what scaffold iteration adds on top of that baseline, and the weight update trajectory traces what weight updates add on top of the harness-only best. We treat this as our primary baseline structure. Across all tasks, the Feedback-Agent begins with scaffold iteration and switches to weight updates once harness progress stalls; we report \textbf{SIA-H} (harness-only best) and \textbf{SIA-W+H} (harness + weight updates best) to isolate each lever's contribution.

\subsection{Per-task results.}\label{sec:per-task}

\subsubsection{LawBench: 191-Class Chinese Criminal Charge Classification.}

LawBench (\citelink{fei2023}{Fei et al., 2023}) is a multi-class legal document classification benchmark drawn from real Chinese criminal case descriptions. Given a factual case summary, the model must identify the correct criminal charge from 191 distinct categories in Chinese statutory law. The 191 classes encode fine-grained legal distinctions that even trained practitioners find demanding: categories of theft (ordinary theft, public-property theft, embezzlement), assault (simple, aggravated, grievous bodily harm), and fraud variants each differ in legally precise factual elements with direct consequences for sentencing. A random-guess baseline is correct less than one percent of the time. The benchmark contains 5,332 training samples and 913 test samples; all evaluations are on the held-out test split.

\emph{Harness updates.} Early scaffold iterations established a working classification pipeline; subsequent generations restructured it around a TF-IDF\,+\,LinearSVC pipeline, iteratively tuning the character $n$-gram range and regulariser $C$, steadily improving accuracy until gains levelled off at \textbf{50.0\%}, a \textbf{36.5 percentage point gain} over the initial run. At this point the Feedback-Agent detected stalling reward and switched to weight updates.

\begin{figure}[H]
\centering
\includegraphics[width=0.60\linewidth]{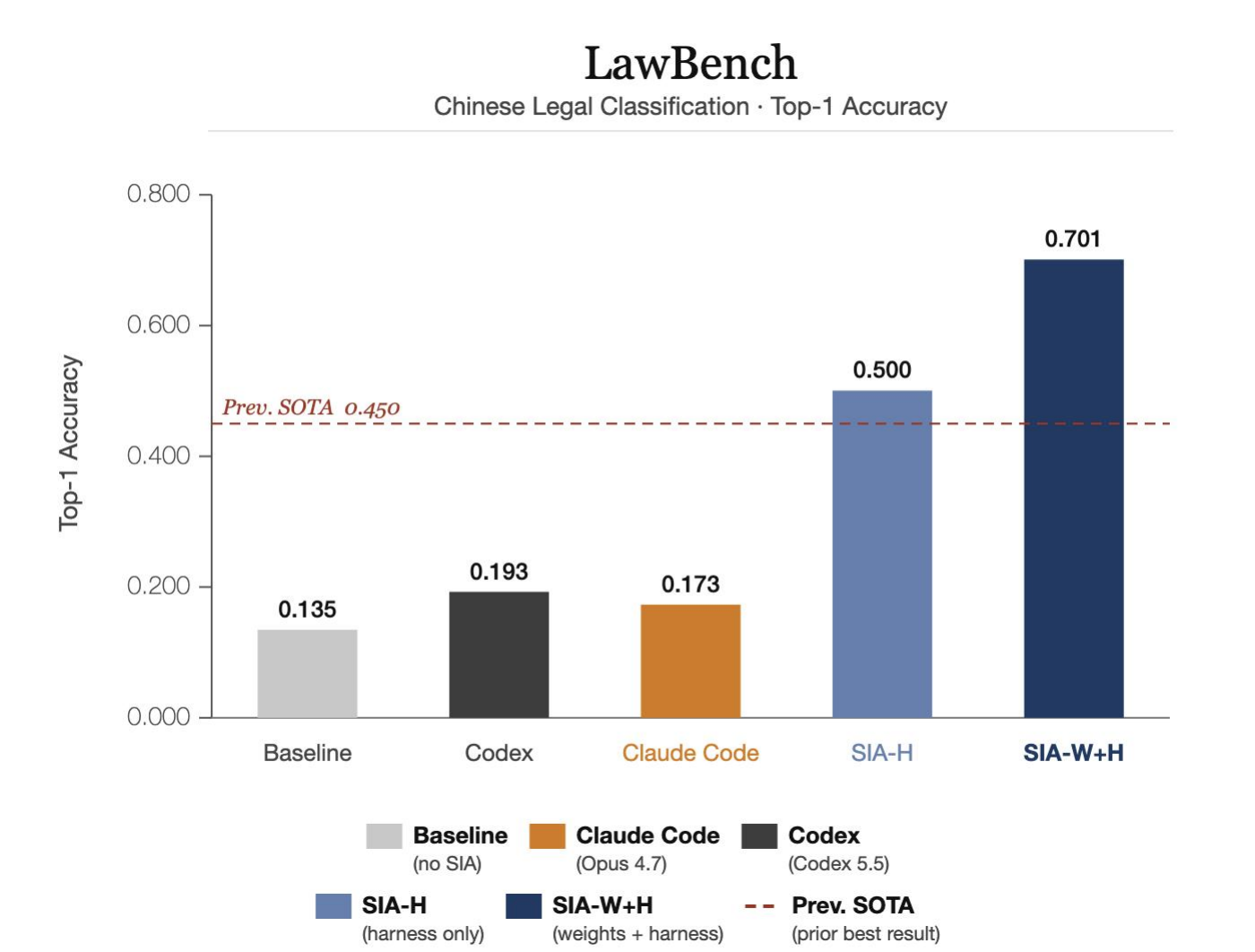}
\caption{\textbf{LawBench results.} Top-1 accuracy for Baseline, SIA-H (harness only), and SIA-W+H (harness + weight updates). Dashed line: prior state-of-the-art.}
\label{fig:3}
\end{figure}

\emph{Weight updates.} Because the reward signal is a clean outcome-based scalar (correct charge or not) and rollouts — each a generated solution script executed against the test split — are cheap to generate in parallel, the Feedback-Agent applied PPO with GAE: a learned value head $V_\phi$ produces per-token advantage estimates over the generated solution script, and a clipped surrogate objective $\min(r_t \hat{A}_t,\,\text{clip}(r_t,1\pm\varepsilon)\hat{A}_t)$ prevents the policy from drifting far from a working baseline. This applied stable gradient pressure on the model's ability to write classification code that correctly handles fine-grained charge distinctions, pushing accuracy to \textbf{70.1\%}, an additional \textbf{20.1 percentage point gain} over the harness-only best (Figure~\ref{fig:3}).

\subsubsection{AlphaEvolve TriMul: CUDA Kernel Optimisation for Protein Structure Prediction.}

The triangular multiplicative update (TriMul) is a core operation in AlphaFold2's Evoformer module, used to propagate pairwise residue-interaction features during protein structure prediction. The task, drawn from the AlphaEvolve benchmark, asks an agent to write a custom CUDA kernel for this operation on an H100 GPU. TriMul is memory-bandwidth-limited rather than compute-limited: threads access non-contiguous memory due to the triangular sparsity structure, inducing warp divergence and cache misses that defeat standard dense-matrix optimisation techniques. Achieving high throughput requires H100-specific knowledge, tensor core scheduling, shared-memory tiling, register pressure management, that standard libraries (cuBLAS, cuSPARSE) do not apply to this operation. Score is defined as $1500 / \text{runtime}$, so a higher score means a faster kernel.

\emph{Harness updates.} The agent progressively built and refined working CUDA kernels across iterations, converging on a best runtime of \textbf{12,483}, a \textbf{1.14$\times$} speedup. Incremental scaffold changes (memory layout hints, compilation flags, retry logic) continued to yield smaller gains until the trajectory plateaued, at which point the Feedback-Agent switched to weight updates.

\begin{figure}[H]
\centering
\includegraphics[width=0.60\linewidth]{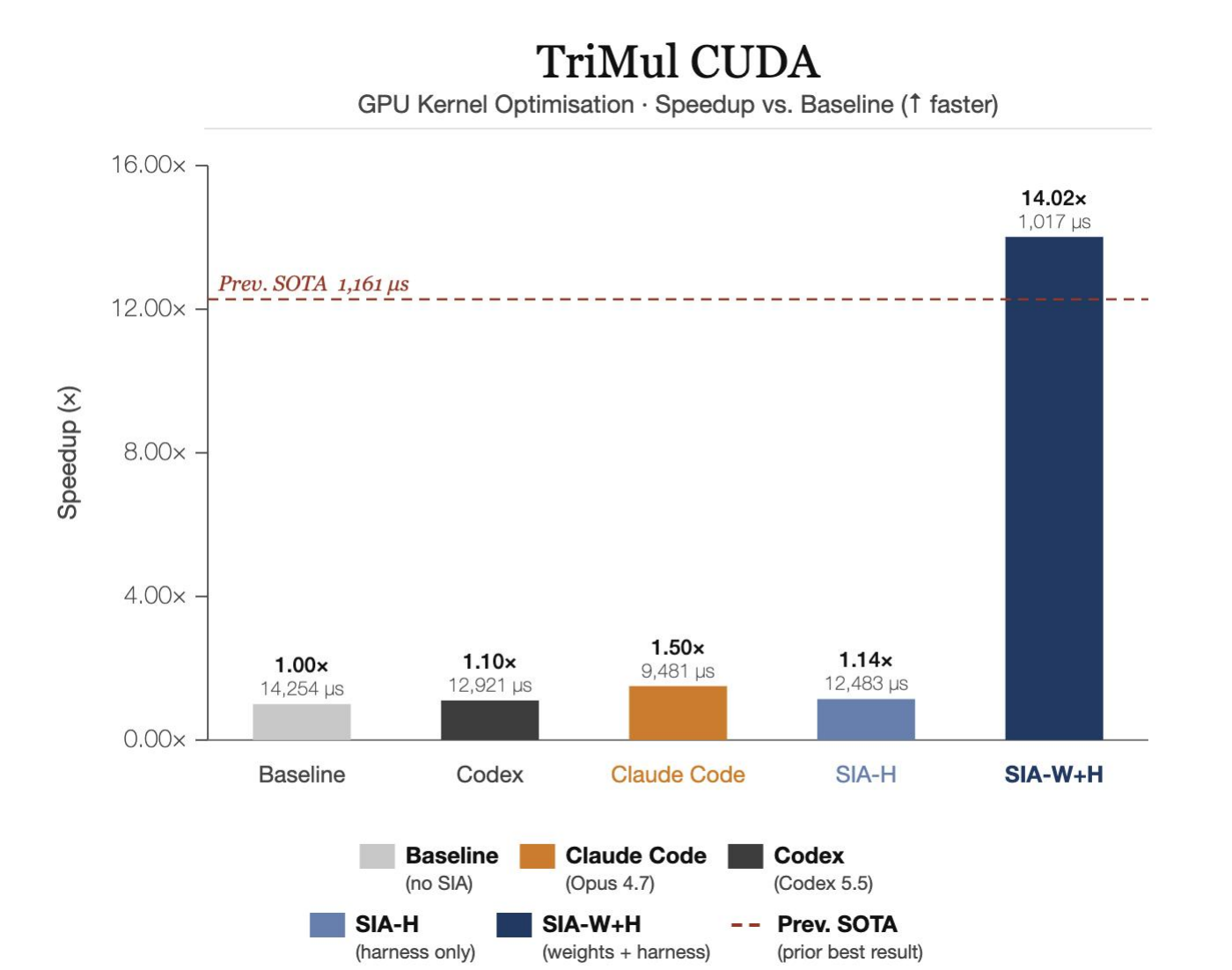}
\caption{\textbf{TriMul CUDA results.} Speedup over baseline for Baseline, SIA-H (harness only), and SIA-W+H (harness + weight updates). Dashed line: prior state-of-the-art.}
\label{fig:4}
\end{figure}

\emph{Weight updates.} Kernel optimisation has a sparse, outcome-heavy reward structure: most generated kernels either fail to compile or are far from optimal, making raw gradient signal from a cold start uninformative. The Feedback-Agent applied entropic advantage weighting, which up-weights high-reward rollouts and discounts near-zero-reward noise via softmax redistribution with adaptive temperature $\beta$: $w_i \propto \exp(r_i/\beta)$, enabling productive gradient flow even when most kernels in a rollout batch are poor. This allowed the model to internalise H100-specific design patterns, shared-memory tiling, fp32 register accumulation, block-size selection, that no scaffold edit could encode, driving runtime down to \textbf{1,017} and a final speedup of \textbf{14.02$\times$}, a \textbf{91.9\% reduction} from the harness-only peak (Figure~\ref{fig:4}).

\subsubsection{MAGIC scRNA-seq Denoising: Single-Cell RNA Imputation.}

Single-cell RNA sequencing (scRNA-seq) measures gene expression across thousands of individual cells, but the resulting count matrices are highly sparse: many true non-zero counts are observed as zero due to technical dropout. MAGIC (Markov Affinity-based Graph Imputation of Cells) addresses this by constructing a $k$-nearest-neighbour graph over cells, computing Markov transition probabilities, and diffusing expression values across graph neighbours to impute missing signal. The task asks an agent to tune MAGIC's coupled hyperparameters, number of neighbours $k$, diffusion steps $t$, kernel bandwidth $\alpha$, and preprocessing choices, on pancreas scRNA-seq data (\citelink{baron2016}{Baron et al., 2016}). The optimisation is non-trivial: $k$ too small overfits to individual cell noise; too large causes over-smoothing that destroys true biological signal. Evaluation uses \texttt{mse\_norm}, a normalised reconstruction quality score against ground truth (higher is better; 1.0 is perfect imputation).

\emph{Harness updates.} The agent swept the coupled hyperparameter space of MAGIC, neighbours $k$, diffusion steps $t$, bandwidth $\alpha$, across several iterations and reached a stable plateau, with \texttt{mse\_norm} settling at a best of \textbf{0.241}. Further scaffold iterations produced no meaningful improvement, prompting the Feedback-Agent to switch to weight updates.

\begin{figure}[H]
\centering
\includegraphics[width=0.60\linewidth]{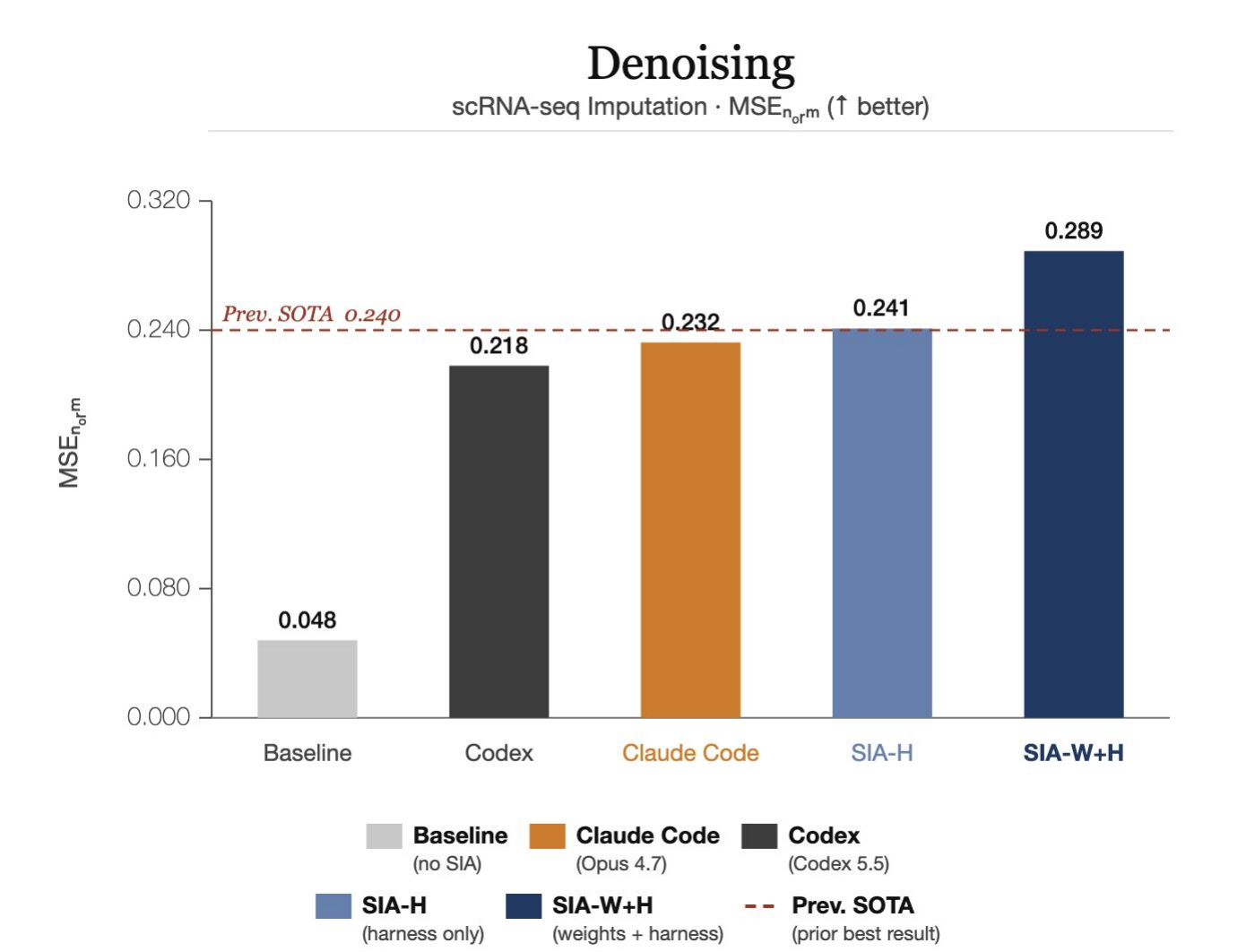}
\caption{\textbf{Denoising results.} MSE$_\text{norm}$ for Baseline, SIA-H (harness only), and SIA-W+H (harness + weight updates). Dashed line: prior state-of-the-art.}
\label{fig:5}
\end{figure}

\emph{Weight updates.} Using GRPO, the model moved beyond parameter tuning entirely. Crucially, the first weight-update checkpoint introduced a structural transformation that the scaffold-only loop, across all harness iterations, never generated: a two-line post-processing step (\texttt{np.clip\,+\,np.rint}) that rounds imputed counts to non-negative integers, enforcing a biological invariant that is trivially correct yet absent from any prior scaffold version. This lifted \texttt{mse\_norm} to \textbf{0.289}, a \textbf{20\% gain} over the harness-only best (Figure~\ref{fig:5}).

\section{Discussion}

\subsection{Combined vs.\ harness-only (RQ1)}\label{sec:combined-vs-harness}

To isolate each lever's contribution we ablate SIA-H (harness updates only) against SIA-W+H (harness + weight updates). Table~\ref{tab:ablation} reports the initial score, prior SOTA, and both operating points across all three tasks.

\begin{table}[H]
\centering
\captionsetup{font=small,skip=6pt,labelfont={bf,sf},labelsep=period}
\caption{\textbf{Ablation: SIA-H vs.\ SIA-W+H.} ``Initial'' is the \texttt{gpt-oss-120b} score through the Meta-Agent's initial scaffold $A_1$. SIA-H is the harness-only best; SIA-W+H adds weight updates.}
\label{tab:ablation}
\footnotesize
\setlength{\tabcolsep}{5pt}
\renewcommand{\arraystretch}{1.15}
\begin{tabularx}{\linewidth}{@{}lcccc@{}}
\toprule
\textbf{Task} & \textbf{Initial} & \textbf{Prev.\ SOTA} & \textbf{SIA-H (harness only)} & \cellcolor{lavrow}\textbf{SIA-W+H (harness + weights)} \\
\midrule
LawBench (top-1 acc) & 13.5\% & 45.0\% & 50.0\% & \cellcolor{lavrow}\textbf{70.1\%} \\
AlphaEvolve TriMul (reward) & 0.105 & 1.292 & 0.120 & \cellcolor{lavrow}\textbf{1.475} \\
Denoising (\texttt{mse\_norm}) & 0.048 & 0.240 & 0.241 & \cellcolor{lavrow}\textbf{0.289} \\
\bottomrule
\end{tabularx}
\end{table}

SIA-W+H strictly outperforms SIA-H on every task, confirming RQ1. The gains are substantial: \textbf{+20.1 pp} on LawBench, \textbf{91.9\%} runtime reduction on TriMul (12,483 $\to$ 1,017 $\mu$s), and \textbf{20\%} on denoising. Each lever occupies a distinct change space, external scaffold versus internal parameters, so neither saturates the gain available from the other (see \S\ref{sec:harness-change}--\ref{sec:weight-change}).

\subsection{What does harness iteration change? (RQ2a)}\label{sec:harness-change}

Harness iteration produces \emph{externalised} changes, new tools, tighter parsers, search procedures, retry policies, and prompt structure, while model weights stay fixed.

Across the three tasks, the Feedback-Agent was observed building increasingly specialised scaffolding: on LawBench, a structured answer-extraction layer and an SVC re-ranker over the model's top candidates; on TriMul, a compilation-error parser that fed CUDA diagnostics back as structured context and a timing harness returning median runtime; on MAGIC denoising, a batched configuration driver and a result-parsing tool that organised (parameter-set, score) pairs for the model to reason over.

In all three cases, the changes are software-engineering improvements: new tools, tighter output parsers, smarter retry logic. The model checkpoint is unchanged throughout; all gains come from how the scaffold mediates between the model and the task environment.

\subsection{How the Feedback-Agent applies weight updates (RQ2b)}\label{sec:weight-how}

The Feedback-Agent does not run a fixed RL procedure. While this paper reports three tasks, we have been running SIA across a wider set of tasks not included here; the algorithm descriptions below reflect common patterns observed across that broader experimentation. In the three tasks reported, PPO with GAE was observed on LawBench; entropic advantage weighting on TriMul; and GRPO on denoising. A fuller treatment of algorithm selection across tasks is deferred to v2.

\begin{itemize}
    \item \textbf{PPO with GAE.} \emph{Observed when:} step-level rewards are dense and training stability is the binding constraint, multi-step tool-use or long code-generation tasks where a single catastrophic update would collapse the policy. A learned value head $V_\phi$ produces per-token advantage estimates $\hat{A}_t = \sum_l (\gamma\lambda)^l \delta_{t+l}$; a clipped surrogate $\min(r_t \hat{A}_t,\,\text{clip}(r_t,1\pm\varepsilon)\hat{A}_t)$ prevents the policy from leaving the trust region. The dual actor-critic optimisation is expensive but yields the lowest-variance gradient signal available.

    \item \textbf{GRPO.} \emph{Observed when:} rollouts are cheap to sample and the verifier fires at episode end, classification, short-answer, or unit-test tasks where hundreds of completions can be scored in a single forward pass. Advantages are normalised within a rollout group of size $G$: $\hat{A}_i = (r_i - \bar{r}) / \sigma_r$, eliminating the value network entirely. This halves memory and enables large parallel batches.

    \item \textbf{Entropic advantage weighting.} \emph{Observed when:} the reward histogram is heavily right-skewed, tasks where correct solutions are rare but individually high-signal, such as hard mathematical proofs or low-pass-rate code synthesis. Rather than zeroing out below-average rollouts, gradient mass is redistributed via softmax with adaptive temperature $\beta$: $w_i \propto \exp(r_i/\beta)$ (\citelink{yuksek2026}{Yuksekgonul et al., 2026}). The temperature is tuned online so that the effective sample size stays above a floor threshold, preventing collapse onto a single trajectory.

    \item \textbf{REINFORCE + KL-to-base.} \emph{Observed when:} the reward is dense and the primary risk is capability regression rather than gradient variance, fine-grained domain-adaptation tasks where the base model is already near-capable and large parameter movement is undesirable. Monte Carlo returns $R_t = \sum_{t'\geq t} \gamma^{t'-t} r_{t'}$ serve as advantages directly, augmented with a penalty $\alpha\,\mathrm{KL}(\pi_\theta \| \pi_{\theta_0})$ against the frozen reference. No critic, no grouping, the simplest possible training loop.

    \item \textbf{Best-of-$N$ behavioural cloning.} \emph{Observed when:} reward is so sparse that $\mathbb{E}[r] \approx 0$ across all rollouts and policy gradient signal is numerically zero. The Feedback-Agent invokes this as a phase-zero cold-start: the top-$k$ rollouts by verifier score are distilled into the model via cross-entropy loss, raising the baseline pass rate to a level where a subsequent PPO or GRPO phase becomes viable.

    \item \textbf{DPO.} \emph{Observed when:} the verifier can rank outputs but not score them absolutely, tasks with soft quality criteria where ordinal signal is reliable but cardinal reward is not. Given a winning rollout $y^+$ and a losing rollout $y^-$, the objective $-\log\sigma\!\left(\beta\log\frac{\pi_\theta(y^+)}{\pi_{\theta_0}(y^+)} - \beta\log\frac{\pi_\theta(y^-)}{\pi_{\theta_0}(y^-)}\right)$ is minimised directly without a reward model.
\end{itemize}

Each training approach is conditioned on trajectory observations rather than a fixed schedule.

\subsection{What weight updates change (RQ2b)}\label{sec:weight-change}

Weight updates produce \emph{internalised} knowledge: domain-specific patterns encoded into the model's parameters that no scaffold edit reaches. Unlike harness changes, which modify the infrastructure surrounding the model, weight updates modify the model's prior over solutions directly.

On LawBench, gradient pressure on the 191-class charge taxonomy sharpened the model's disambiguation of adjacent categories, distinguishing theft sub-types, assault grades, and fraud variants, without any prompt-side hint. On AlphaEvolve TriMul, the weights converged on H100-specific kernel design patterns (shared-memory tiling, fp32 register accumulation, block-size selection) that the base model never produced regardless of scaffold quality. On MAGIC denoising, the first weight-update checkpoint introduced a structural invariant the harness had never proposed: a \texttt{np.clip\,+\,np.rint} post-processing step that rounded imputed counts to non-negative integers, encoding a biological constraint directly into the policy.

In each case the internalised knowledge is task-specific and verifier-aligned, it emerges from direct gradient pressure, not from any human-authored instruction. The harness shapes \emph{how} the agent searches; weight updates change \emph{what} the model knows.

\section{Limitations}

\textbf{Coupled co-evolutionary Goodhart.}
Harness search and RL weight updates both optimise against the same fixed verifier $V$.
Each pass shapes the distribution the other sees: the harness finds scaffolds that are easy for the current policy to exploit; the weights train on data collected through a scaffold that will subsequently change.
The joint fixed point of this coupled system is a Nash equilibrium between two optimisers that are blind to each other's update history, not a point that maximises $V$ on out-of-distribution scaffolds or novel policies.
Standard Goodhart analyses assume a single optimiser; the two-lever setting produces a \emph{coupled} variant whose fixed points can appear strong on the training verifier while being fragile under any perturbation to either component.

\section{Future Work}

\textbf{Meta-RL over the action-selection policy.}
The Feedback-Agent currently selects between harness and weight updates using a frozen LLM prior.
A more principled approach treats the selection policy itself as the object to be learned: run SIA across a distribution of tasks, treat each (trajectory, action, outcome) triple as a transition in an outer MDP, and train the selector via RL on that outer MDP.
The selector then improves its lever-attribution through experience across tasks rather than relying on fixed heuristics calibrated on trajectories from a different capability regime.
This creates a genuinely recursive structure, a self-improving system whose improvement mechanism is itself self-improving, and raises non-trivial questions about the stability of such nested loops that are distinct from any question in single-level RL or meta-learning.

\textbf{More interleaved training and harness switching.}
The current SIA loop alternates between harness search and weight update phases in discrete, coarse-grained rounds.
A finer-grained schedule, where the Feedback-Agent can trigger a weight update mid-harness search, or resume harness exploration immediately after a gradient step, could reduce the lag between observing a plateau and acting on it, and may unlock improvement trajectories that coarse alternation misses.

\bibliographystyle{plain}
\bibliography{verified_refs}

\end{document}